\documentclass[11pt,a4paper]{article}
\makeatletter
\providecommand{\@LN}[2]{}
\makeatother
\usepackage[hyperref]{acl2021}
\usepackage{times}
\usepackage{latexsym}
\usepackage{graphicx}
\usepackage{booktabs}
\usepackage{multirow}
\usepackage{inconsolata}
\usepackage{xspace}
\usepackage{subcaption}
\usepackage{amsmath}
\usepackage{cleveref}
\usepackage{paralist}
\usepackage{enumitem}

\usepackage{microtype}

\aclfinalcopy 
\def\aclpaperid{1253} 

\newcommand{\eat}[1]{}

\title{
  Towards Robustness of Text-to-SQL Models against Synonym Substitution
  }

\author{Yujian Gan${}^{1}$ \ \ \ \ Xinyun Chen${}^{2}$ \ \ \ \ Qiuping Huang${}^{4}$ \ \ \ \  Matthew Purver${}^{1,3}$  \\ \bf  John R. Woodward${}^{1}$ \ \ \ \ Jinxia Xie${}^{5}$ \ \ \ \ Pengsheng Huang${}^{6}$ \\

${}^{1}$Queen Mary University of London \ \ \ \  \ \ \ \ ${}^{2}$UC Berkeley
\ \ \  \ \ \ \ ${}^{3}$Jožef Stefan Institute\\
${}^{4}$Nanning Central Sub-branch of the People's Bank of China \\
${}^{5}$Guangxi University of Finance and Economics  \ \ \ \  \ \ \ \  ${}^{6}$Beijing MeiyiLab Co.,Ltd. \\
  \texttt{\{y.gan,m.purver,j.woodward\}@qmul.ac.uk}  \ \ \ \  \ \ \ \  \texttt{xinyun.chen@berkeley.edu} \\
 \texttt{qiuping\_h@foxmail.com} \ \  \ \   \texttt{jinxia\_xie@hotmail.com}  \ \  \ \   \texttt{huangpengsheng@pku.edu.cn} \\

  }

\date{}

\begin{document}
\maketitle
\begin{abstract}
  Recently, there has been significant progress in studying neural networks to translate text descriptions into SQL queries.
  Despite achieving good performance on some public benchmarks, existing text-to-SQL models typically rely on the lexical matching between words in natural language (NL) questions and tokens in table schemas, which may render the models vulnerable to attacks that break the schema linking mechanism.
  In this work, we investigate the robustness of text-to-SQL models to synonym substitution. In particular, we introduce Spider-Syn, a human-curated dataset based on the Spider benchmark for text-to-SQL translation. NL questions in Spider-Syn are modified from Spider, by replacing their schema-related words with manually selected synonyms that reflect real-world question paraphrases. We observe that the accuracy dramatically drops by eliminating such explicit correspondence between NL questions and table schemas, even if the synonyms are not adversarially selected to conduct worst-case adversarial attacks~\footnote{Following the prior work on adversarial learning, worst-case adversarial attacks mean adversarial examples generated by attacking specific models.}.
  Finally, we present two categories of approaches to improve the model robustness.
  The first category of approaches utilizes additional synonym annotations for table schemas by modifying the model input, while the second category is based on adversarial training. We demonstrate that both categories of approaches significantly outperform their counterparts without the defense, and the first category of approaches are more effective.~\footnote{Our code and dataset is available at  \href{https://github.com/ygan/Spider-Syn}{https://github.com/ygan/Spider-Syn}
  }

\end{abstract}

\section{Introduction}

Neural networks have become the defacto approach for various natural language processing tasks, including text-to-SQL translation. Various benchmarks have been proposed for this task, including earlier small-scale single-domain datasets such as ATIS and GeoQuery~\cite{,data-sql-imdb-yelp,data-atis-geography-scholar,data-geography-original}, and recent large-scale cross-domain datasets such as WikiSQL~\cite{zhongSeq2SQL2017} and Spider~\cite{Yu2018a}. While WikiSQL only contains simple SQL queries executed on single tables, Spider covers more complex SQL structures, e.g., joining of multiple tables and nested queries.

\begin{figure}[t]
    \includegraphics[width=0.47\textwidth]{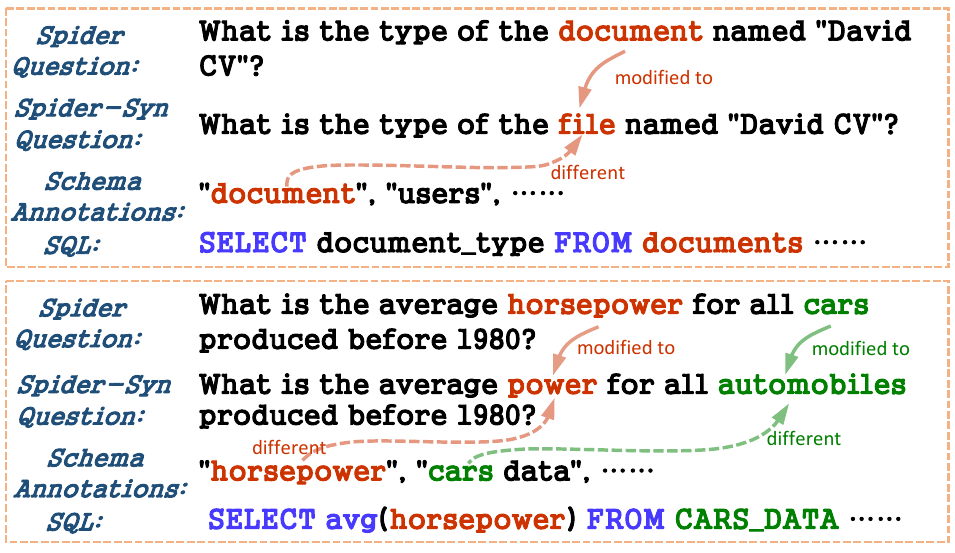}
    \centering
    \caption{Sample Spider questions that include the same tokens as the table schema annotations, and such questions constitute the majority of the Spider benchmark. In our Spider-Syn benchmark, we replace some schema words in the NL question with their synonyms, without changing the SQL query to synthesize.}
    \label{figure:spider-syn}
  \end{figure}

\eat{
\begin{figure}[t]
    \includegraphics[width=0.47\textwidth]{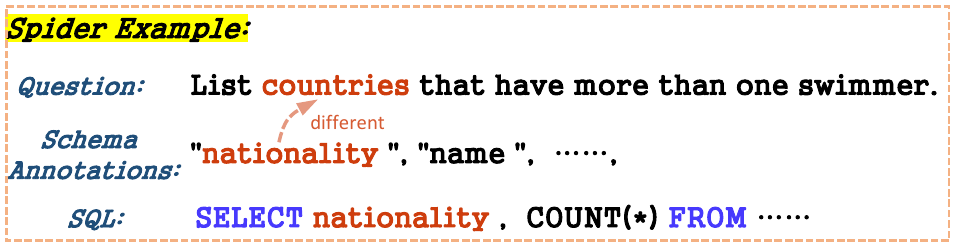}
    \centering
    \caption{
      Questions of not using schema annotations in Spider, we name it synonym substitution questions.
       }
    \label{figure:original-example}
  \end{figure}
}

The state-of-the-art models have achieved impressive performance on text-to-SQL tasks, e.g., around 70\% accuracy on the Spider test set, even if the model is tested on databases that are unseen in training.
However, we suspect that such cross-domain generalization heavily relies on the exact lexical matching between the NL question and the table schema. 
As shown in Figure~\ref{figure:spider-syn}, names of tables and columns in the SQL query are explicitly stated in the NL question. Such questions constitute the majority of cross-domain text-to-SQL benchmarks including both Spider and WikiSQL. Although assuming 
exact lexical matching is a good starting point to solving the text-to-SQL problem, this assumption usually does not hold in real-world scenarios. Specifically, it requires that users have 
precise knowledge of the table schemas to be included in the SQL query, which could be tedious for synthesizing complex SQL queries.

\eat{Although the Spider and WikiSQL contain some examples that use synonyms instead of schema annotations, as shown in Figure \ref{figure:original-example}, these examples account for only a small proportion.}

In this work, we investigate whether 
state-of-the-art text-to-SQL models preserve good prediction performance without the assumption of 
exact lexical matching, where NL questions use synonyms to refer to tables or columns in SQL queries. We call such NL questions~\emph{synonym substitution} questions.
Although some existing approaches can automatically generate synonymous substitution examples, these examples may deviate from real-world scenarios, e.g., they may not follow common human writing styles, or even accidentally becomes inconsistent with the annotated SQL query.
To provide a reliable benchmark for evaluating 
model performance on synonym substitution questions, we introduce Spider-Syn, a  human-curated dataset constructed by modifying NL questions in the Spider dataset.
Specifically, we replace the schema annotations in the NL question with 
synonyms, manually selected so as not to change the corresponding SQL query, as shown in Figure \ref{figure:spider-syn}.
We demonstrate that when models are only trained on the original Spider dataset, they suffer a significant performance drop on Spider-Syn, even though the Spider-Syn benchmark is not constructed to exploit the worst-case attacks for text-to-SQL models. 
It is therefore clear that the performance of these models will suffer in real-world use, particularly in cross-domain scenarios.

To improve the robustness of text-to-SQL models, we utilize synonyms of table schema words, which are either manually annotated, or automatically generated when no annotation is available. We investigate two categories of approaches to incorporate these synonyms. The first category of approaches modify the schema annotations of the model input, so that they align better with the NL question. No additional training is required for these approaches.
The second category of approaches are based on adversarial training, where we augment the training set with NL questions modified by synonym substitution.
Both categories of approaches significantly improve the robustness, and the first category is both effective and requires less computational resources.

In short, we make the following contributions: 

\begin{itemize}[leftmargin=*,noitemsep,topsep=0em]
    \item We conduct a comprehensive study to evaluate the robustness of text-to-SQL models against synonym substitution.
    \item Besides worst-case adversarial attacks, we further introduce Spider-Syn, a human-curated dataset built upon Spider, to evaluate synonym substitution for real-world question paraphrases. 
    \item We propose a simple yet effective approach to utilize multiple schema annotations, without the need of additional training. We show that our approach outperforms adversarial training methods on Spider-Syn, and achieves competitive performance on worst-case adversarial attacks.
\end{itemize}
\section{Spider-Syn Dataset}

\begin{figure}[t]
    \includegraphics[width=0.47\textwidth]{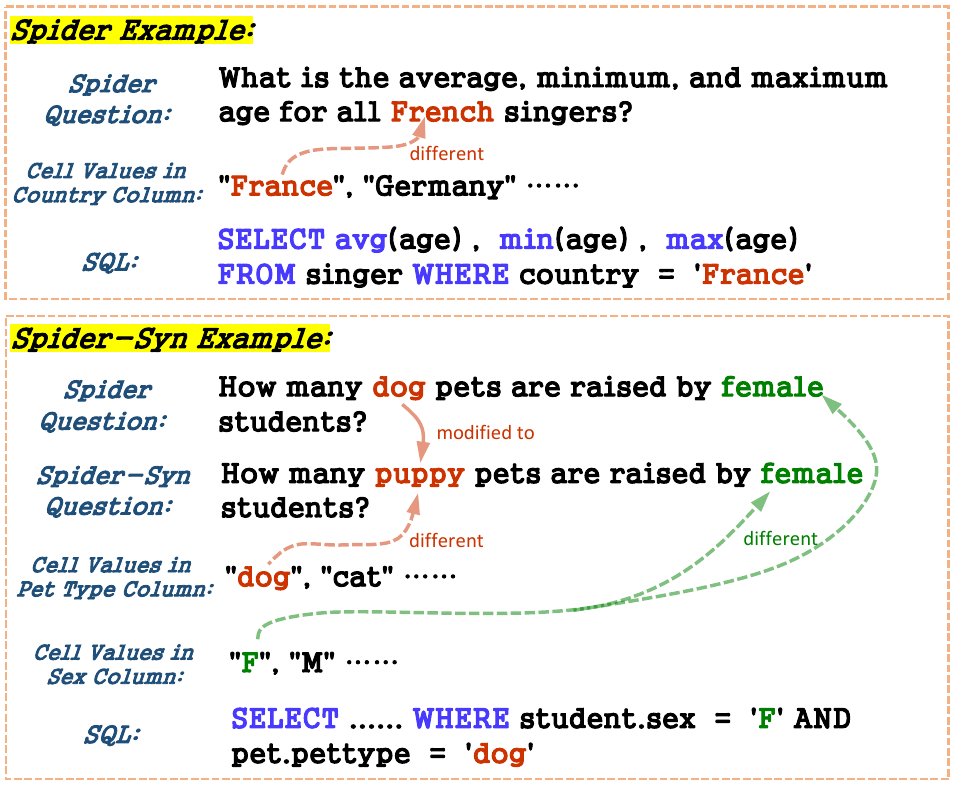}
    \centering
    \caption{Synonym substitution  occurs in cell value words in both Spider and Spider-Syn.}
    \label{figure:cell-value}
  \end{figure}

\subsection{Overview}
\label{section:2Overall}
We construct the Spider-Syn benchmark by manually modifying NL questions in the Spider dataset using synonym substitution.
The purpose of building Spider-Syn is to simulate the scenario where users do not call the exact schema words in the utterances, e.g., users may not have the knowledge of table schemas.
In particular, we focus on synonym substitution for words related to databases, including table schemas and cell values.
Consistent with Spider, Spider-Syn contains 7000 training and 1034 development examples, but Spider-Syn does not contain a test set since the Spider test set is not public. 
Figure \ref{figure:spider-syn} presents two examples in Spider-Syn and how they are modified from Spider.

\subsection{Conduct Principle}
\label{sec:syn-principle}

The goal of constructing the Spider-Syn dataset is not to perform worst-case adversarial attacks against existing text-to-SQL models, but to investigate the model robustness for paraphrasing schema-related words, which is particularly important when users do not have the knowledge of table schemas.
We carefully select the synonyms to replace the original text to ensure that new words will not cause ambiguity in some domains. For example, the word `\emph{country}' can often be used to replace the word `\emph{nationality}'. 
However, we did not replace it in the domain whose `\emph{country}' means people's `\emph{born country}' different from its other schema item, `\emph{nationality}'.
Besides, some synonym substitutions are only valid in the specific domain. 
For example, the word `\emph{number}' and `\emph{code}' are not generally synonymous, but `\emph{flight number}' can be replaced by `\emph{flight code}' in the aviation domain.

\begin{figure}[t]
  \includegraphics[width=0.47\textwidth]{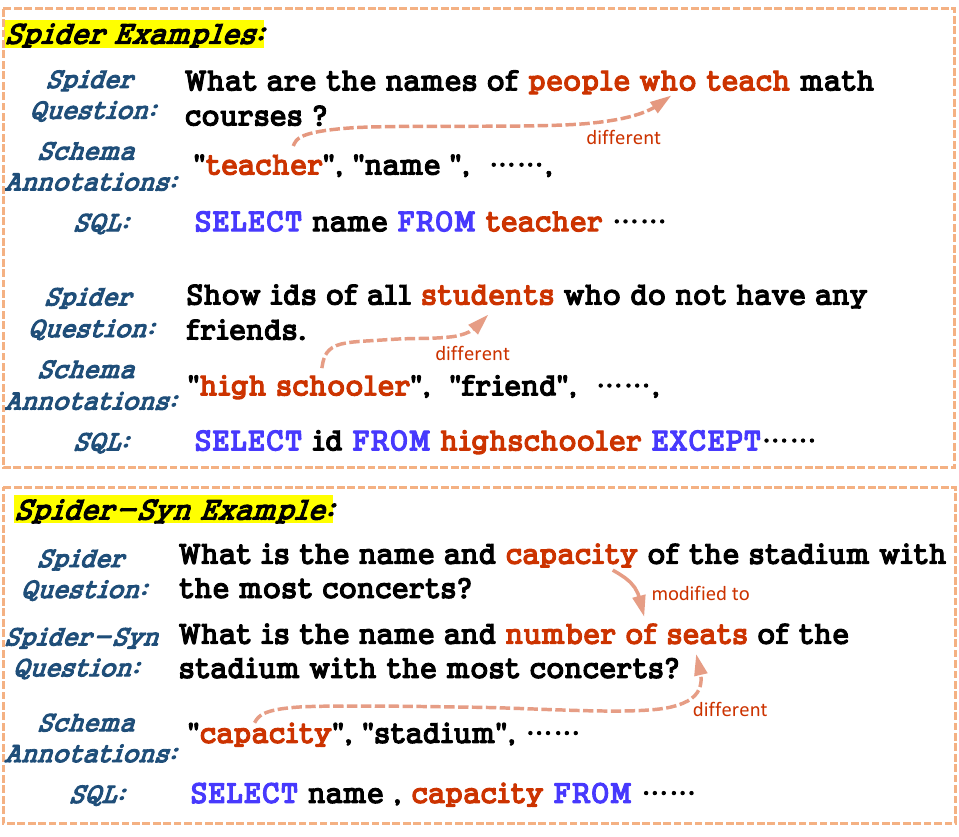}
  \centering
  \caption{Samples of replacing the original words or phrases by synonymous phrases.}
  \label{figure:syn-exp}
\end{figure}

Most synonym substitutions use 
relatively common words\footnote{According to 20,000 most common English words in~\url{https://github.com/first20hours/google-10000-english}.} to replace the schema item words. 
Besides, we denote `\emph{id}', `\emph{age}', `\emph{name}', and `\emph{year}' as reserved words, which are the most standard words to represent their meanings. Under this principle, we keep some original Spider examples unchanged in Spider-Syn.
Our synonym substitution does not guarantee that the modified NL question has the exact same meaning as the original question, but guarantees that its corresponding SQL is consistent.
In Figure~\ref{figure:cell-value}, Spider-Syn replaces the cell value word `\emph{dog}' with `\emph{puppy}'. Although puppy is only a subset of dog, the corresponding SQL for the Spider-Syn question should still use the word `\emph{dog}' instead of the word `\emph{puppy}' because there is only dog type in the database and no puppy type. 
Similar reasoning is needed to infer that the word `\emph{female}' corresponds to `\emph{F}' in Figure \ref{figure:cell-value}.

In some cases, words are replaced by synonymous phrases (rather than single words), as shown in Figure~\ref{figure:syn-exp}.
Besides, some substitutions are also based on the database contents.
For example, a column `\emph{location}' of the database `\emph{employee\_hire\_evaluation}' in Spider only stores city names as cell values.
Without knowing the table schema, users are more likely to call `\emph{city}' instead of `\emph{location}' in their NL questions.

To summarize, we construct Spider-Syn with the following principles: 

\begin{itemize}[leftmargin=*,noitemsep,topsep=0em]
    \item 
    Spider-Syn is not constructed to exploit the worst-case adversarial attacks, but to represent real-world use scenarios; it therefore uses only relatively common words as substitutions.
    \item We conduct synonym substitution only for words related to schema items and cell values.
    \item Synonym substitution includes both single words and phrases with multiple words.
\end{itemize}

\begin{figure}[t]
  \includegraphics[width=0.47\textwidth]{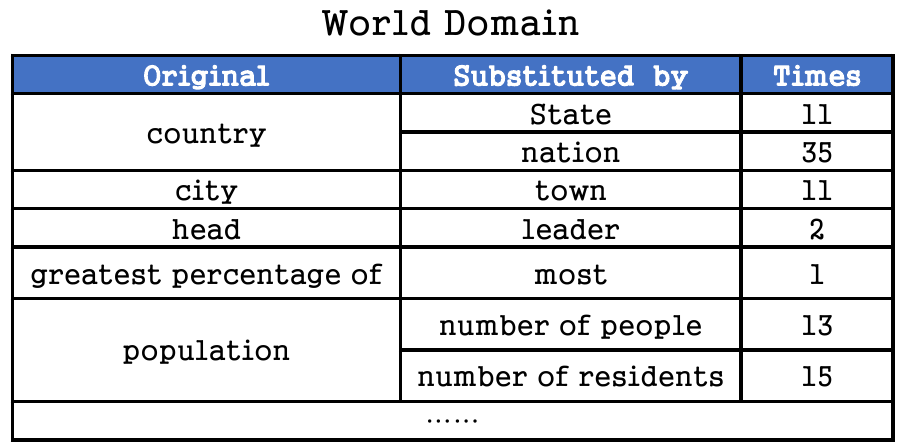}
  \centering
  \caption{Examples of synonym substitutions in the `world' domain from Spider-Syn.}
  \label{figure:report-world}
\end{figure}

\subsection{Annotation Steps}
Before annotation, we first separate original Spider samples based on their domains. For each domain, we only utilize synonyms that are suitable for that domain.
We recruit four graduate students major in computer science to annotate the dataset manually.
They are trained with a detailed annotation guideline, principles, and some samples. 
One is allowed to start after his trial samples are approved by the whole team.

As synonyms can be freely chosen by annotators, standard inter-annotator agreement metrics are not sufficient to confirm the data quality. Instead, 
we conduct the quality control with two rounds of review. 
The first round is the cross-review between annotations. 
We require the annotators to discuss their disagreed annotations and come up with a final result out of consensus.
To improve the work efficiency, we extract all synonym substitutions as a report without the NL questions from the annotated data, as shown in Figure \ref{figure:report-world}. Then, the annotators do not have to go through the NL questions one by one. The second round of review is similar to the first round but is done by native English speakers.

\subsection{Dataset Statistics}
In Spider-Syn, 5672 questions are modified compared to the original Spider dataset.
In 5634 cases the schema item words are modified, with the cell value words modified in only 27 cases.
We use 273 synonymous words and 189 synonymous phrases to replace approximately 492 different words or phrases in these questions.
In all Spider-Syn examples, 
there is an average of 0.997 change per question and 7.7 words or phrases modified per domain.

Besides, Spider-Syn keeps 2201 and 161 original Spider questions in the training and development set, respectively. 
In the modification between the training and development sets, 52 modified words or phrases were the same, accounting for 35\% of the modification in the development set.

\section{Defense Approaches}
We present two categories of approaches for improving model robustness to synonym substitution. 
We first introduce our multiple annotation selection approach, which could utilize multiple annotations for one schema item.
Then we present an adversarial training method based on analysis of the NL question and domain information.

\subsection{Multi-Annotation Selection (MAS)}
\label{chapter:multiple-annotation}


The synonym substitution problem emerges when users do not call the exact names in table schemas to query the database.
Therefore, one defense against synonym substitution is utilizing multiple annotation words to represent the table schema, so that the schema linking mechanism is still effective.
For example, for a database table with the name `\emph{country}', we annotate additional table names with similar meanings, e.g., `\emph{nation}', `\emph{State}', etc. In this way, we explicitly inform the text-to-SQL models that all these words refer to the same table, thus the table should be called in the SQL query when the NL question includes any of the annotated words. 

We design a simple yet effective mechanism to incorporate multiple annotation words, called multiple-annotation selection (MAS).
For each schema item, we check whether any annotations appear in the NL question, and we select such annotations as the model input.
When no annotation appears in the question, we select the default schema annotation, i.e., the same as the original Spider dataset. 
In this way, we could utilize multiple schema annotations simultaneously, without changing the model input format.

The main advantage of this method is that it does not require additional training, and could apply to existing models trained without synonym substitution questions. 
Annotating multiple schema words could be done automatically or manually, and we compare them in Section~\ref{section:experiment}.

\subsection{Adversarial Training}
\label{chapter:adversarial-training}

Motivated by the idea of adversarial training that can improve the robustness of machine learning models against adversarial attacks~\cite{aleks2017deep,Morris2020}, we implement adversarial training using the current open-source SOTA model RAT-SQL \cite{Wang2019}.
We use the BERT-Attack model~\cite{Li2020} to generate adversarial examples, and implement the entire training process based on the TextAttack framework~\cite{Morris2020}. 
Text\-Attack provides 82 pre-trained models, including word-level LSTM, word-level CNN, BERT-Attack, and other pre-trained Transformer-based models.

We follow the standard adversarial training pipeline that iteratively generates adversarial examples, and trains the model on the dataset augmented with these adversarial examples.
When generating adversarial examples for training, we aim to generate samples that align with the Spider-Syn 
principles, rather than arbitrary adversarial perturbations. We describe the details of adversarial example generation below.

\begin{figure*}[t]
    \includegraphics[width=0.95\textwidth]{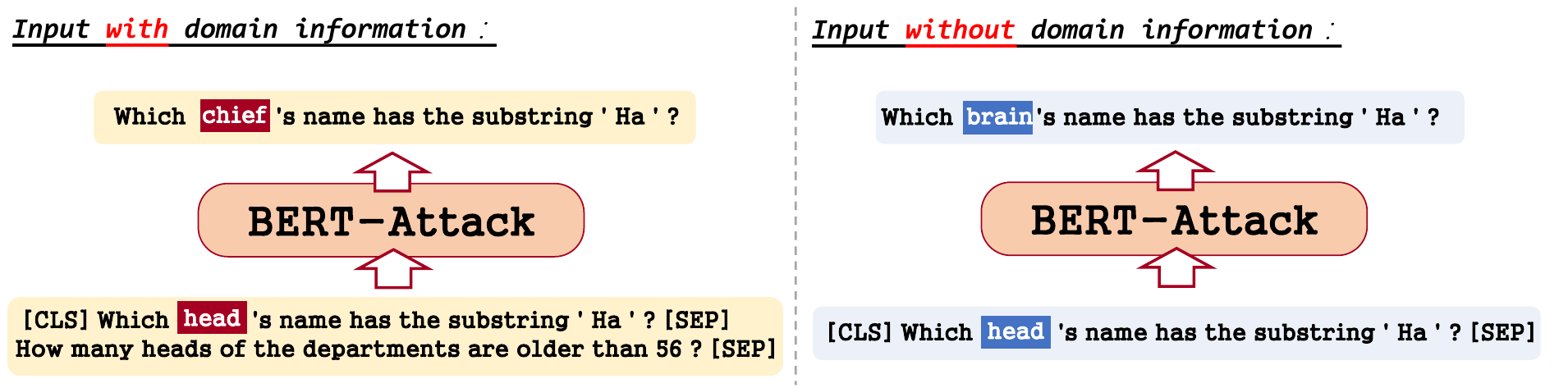}
    \centering
    \caption{Input the BERT-Attack with and without domain information.}
    \label{figure:bert-attack}
  \end{figure*}


\subsubsection{Generating Adversarial Examples}
We choose BERT-Attack to generate the adversarial examples.
Different from other word substitution methods \cite{mrksic-etal-2016-counter,ebrahimi-etal-2018-hotflip,wei-zou-2019-eda}, BERT-Attack model considers the entire NL question when generating words for synonym substitution. 
Such a sentence-based method can generate different synonyms for the same word in different context. 
For example,  the word `\emph{head}' in `\emph{the head of a department}' and `\emph{the head of a body}' should correspond to different synonyms. 
Making such distinctions requires an analysis of the entire sentence, since the keywords' positions may not be close, such as that the word `\emph{head}' and `\emph{department}' are not close in `\emph{Give me the info of heads whose name is Mike in each department}'.

In addition to the original question, we add extra domain information into the BERT-Attack model, as shown in Figure \ref{figure:bert-attack}. 
Without the domain information, on the right side of the Figure \ref{figure:bert-attack}, the BERT-Attack model conjectures the word `\emph{head}' represent the head of a body, since there are multiple feasible interpretations for the word `\emph{head}' if you only look at the question. 
To eliminate the ambiguity, we feed questions with its domain information into the BERT-Attack model, as shown on the left side of the Figure \ref{figure:bert-attack}.

Instead of using schema annotations, we select several other questions from the same domain as domain information. 
These questions should contain the schema item words we plan to replace, and other distinct schema item words in the same domain.
The benefits of using sentences instead of schema annotations as domain information include: 1) avoiding many unrelated schema annotations, which could include hundreds of words; 2) the sentence format is closer to the pre-training data of BERT.
As shown on the left side of the Figure \ref{figure:bert-attack}, our method improves the quality of data generation.


Since our work focuses on the synonym substitution of schema item words, we make two additional constraints to limit the generation of adversarial examples: 1) only words about schema items and cell values can be replaced; and 2) do not replace the reserved words discussed in Section~\ref{sec:syn-principle}.
These constraints make sure that the adversarial examples only perform the synonym substitution for words related to database tables.

\section{Experiments}
\label{section:experiment}
\subsection{Experimental Setup}
We compare our approaches against baseline methods on both the Spider \cite{Yu2018a} and Spider-Syn development sets. As discussed in Section~\ref{section:2Overall}, the Spider test set is not publicly accessible, and thus Spider-Syn does not contain a test set.
Both Spider and Spider-Syn contain 7000 training and 1034 development samples respectively, where there are 146 databases for training and 20 for development.
The SQL queries and schema annotations between Spider and Spider-Syn are the same; 
the difference is that the questions in Spider-Syn are modified from Spider by synonym substitution.
Models are evaluated using the official exact matching accuracy metric of Spider.

We first evaluate open-source models that reach competitive performance on Spider: GNN \cite{Bogin2019}, IRNet \cite{Guo2019} and RAT-SQL \cite{Wang2019}, on the Spider-Syn development set. 
We then evaluate our approaches with RAT-SQL+BERT model (denoted as $\textnormal{RAT-SQL}_{\textnormal{\small{B}}}$) on both Spider-Syn and Spider development set. 

We examine the robustness of following approaches for synonym substitution:

\begin{itemize}[leftmargin=*,noitemsep,topsep=0em]
    \setlength{\parskip}{0pt}
    \item 
    {\bf SPR:} 
    Indicate that the model is trained on the Spider dataset.
    
    \item 
    $\textbf{SPR}_{\bf SYN}\textbf{:}$
    Indicate that the model is trained on the Spider-Syn dataset .
    
    \item 
    $\textbf{SPR}_{\bf SPR\&SYN}\textbf{:}$
    Indicate that the model is trained on both Spider and Spider-Syn datasets.
    
    \item 
    $\textbf{ADV}_{\bf BERT}\textbf{:}$
    To improve the robustness of text-to-SQL models,  we use adversarial training methods to deal with synonym substitution.
    This variant means that we use BERT-Attack following the design introduced in Section \ref{chapter:adversarial-training}. 
    Note that we only use the Spider dataset for adversarial training.

    \item 
    $\textbf{ADV}_{\bf GLOVE}\textbf{:}$
    To demonstrate the effectiveness of our $\textnormal{ADV}_{\textnormal{\small{BERT}}}$ method, we also evaluate a simpler adversarial training method based on the nearest GLOVE word vector \cite{pennington-etal-2014-glove,mrksic-etal-2016-counter}. This method only considers the meaning of a single word, dispensing with domain information and question context.

    \item {\bf ManualMAS:}
    MAS stands for `\emph{multi-annotation selection}', as introduced in Section~\ref{chapter:multiple-annotation}. 
    ManualMAS means that we collect multiple annotations of schema item words, which are synonyms used in Spider-Syn.
    Afterward, MAS selects the appropriate annotation for each schema item as the model input.

    \item {\bf AutoMAS:}
    In contrast to ManualMAS, in AutoMAS we
    collect multiple annotations based on the nearest GLOVE word vector, as used in $\textnormal{ADV}_{\textnormal{\small{GLOVE}}}$. 
    In this way, compared to ManualMAS, there are much more synonyms to be selected from for AutoMAS.
    Both ManualMAS and AutoMAS are to demonstrate the effectiveness of MAS in an ideal case.
This experimental design principle is similar to evaluating adversarially trained models on the same adversarial attack used for training, which aims to show the generalization to in-distribution test samples. 
\end{itemize}

\begin{table}[t]
    \centering
    \resizebox{.99\columnwidth}{!}{
    \smallskip\begin{tabular}{lcc}
        \hline
       \bf model & \bf Spider & \bf Spider-Syn \\
        \hline \hline
        GNN + SPR \cite{Bogin2019}  & 48.5\% & 23.6\% \\ 
        IRNet + SPR \cite{Guo2019}   & 53.2\% & 28.4\% \\ 
        RAT-SQL + SPR \cite{Wang2019}  & 62.7\% &  33.6\% \\ 
        $\textnormal{RAT-SQL}_{\textnormal{\small{B}}}$ + SPR \cite{Wang2019} & 69.7\% &  48.2\% \\ 
        \hline 
    \end{tabular}
    }
    \caption{Exact match accuracy on the Spider and Spider-Syn development set, where models are trained on the original Spider training set.}\smallskip
    \label{table:previous-models}
    
\end{table}

\begin{table}[t]
    \centering
    \resizebox{.99\columnwidth}{!}{
    \smallskip\begin{tabular}{lcc}
        \hline
       \bf SQL Component & \bf Spider & \bf Spider-Syn \\
        \hline \hline
        SELECT   & 0.910 & 0.699 \\ 
        SELECT (no AGG)   & 0.926 & 0.712 \\ 
        WHERE   & 0.772 & 0.715 \\ 
        WHERE (no OP)  & 0.824 & 0.757 \\ 
        GROUP BY (no HAVING)   & 0.846 & 0.575 \\
        GROUP BY   & 0.816 & 0.553 \\
        ORDER BY  & 0.831 &  0.768 \\ 
        AND/OR  & 0.979 &  0.977 \\ 
        IUE  & 0.550 &  0.344 \\ 
        KEYWORDS  & 0.897 &  0.876 \\ 

        \hline 
    \end{tabular}
    }
    \caption{F1 scores of component matching of $\textnormal{RAT-SQL}_{\textnormal{\small{B}}}$+SPR on development sets.}\smallskip
    \label{table:breakdown}
    
\end{table}

\subsection{Results of Models Trained on Spider}

Table \ref{table:previous-models} presents the exact matching accuracy of models trained on the Spider training set, and we evaluate them on development sets of Spider and Spider-Syn. Although Spider-Syn is not designed to exploit the worst-case attacks of text-to-SQL models, compared to Spider, the performance of all models has clearly dropped by about 20\% to 30\% on Spider-Syn. Using BERT for input embedding suffers less performance degradation than models without BERT, but the drop is still significant. These experiments demonstrate that training on Spider alone is insufficient for achieving good performance on synonym substitutions, because the Spider dataset only contains a few questions with synonym substitution.

To obtain a better understanding of prediction results, we compare the F1 scores of $\textnormal{RAT-SQL}_{\textnormal{\small{B}}}$+SPR on different SQL components on both the Spider and Spider-Syn development set.
As shown in Table \ref{table:breakdown}, the performance degradation mainly comes from the components including schema items, while the decline in the `\emph{KEYWORDS}' and the `\emph{AND/OR}' that do not include schema items is marginal. 
This observation is consistent with the design of Spider-Syn, which focuses on the substitution of schema item words.

\begin{table}[t]
    \centering
    \resizebox{.99\columnwidth}{!}{
    \smallskip\begin{tabular}{lcc}
        \hline
       \bf Approach & \bf Spider & \bf Spider-Syn \\
        \hline \hline
        SPR  & \bf 69.7\% &  48.2\% \\ 
        $\textnormal{SPR}_{\textnormal{\small{SYN}}}$  & 67.8\% &  59.9\% \\ 
        $\textnormal{SPR}_{\textnormal{\small{SPR\&SYN}}}$  & 68.1\% &  58.0\% \\ 
        \hline
        $\textnormal{ADV}_{\textnormal{\small{GLOVE}}}$  & 48.7\% &  27.7\% \\ 
        $\textnormal{ADV}_{\textnormal{\small{BERT}}}$  & 68.7\% &  58.5\% \\ 
        \hline
        SPR + ManualMAS  & 67.4\% & \bf 62.6\% \\ 
        SPR + AutoMAS  & 68.7\% &  56.0\% \\ 

        \hline 
    \end{tabular}
    }
    \caption{Exact match accuracy on the Spider and Spider-Syn development set. All approaches use the $\textnormal{RAT-SQL}_{\textnormal{\small{B}}}$ model.}\smallskip
    \label{table:our-approaches}
    
\end{table}

\subsection{Comparison of Different Approaches}

Table \ref{table:our-approaches} presents the results of $\textnormal{RAT-SQL}_{\textnormal{\small{B}}}$ trained with different approaches. 
We focus on $\textnormal{RAT-SQL}_{\textnormal{\small{B}}}$ since it achieves the best performance on both Spider and Spider-Syn, as shown in Table \ref{table:previous-models}. Our MAS approaches significantly improve the performance on Spider-Syn, with only 1-2\% performance degradation on the Spider. With ManualMAS, we see an accuracy of 62.6\%, which outperforms all other approaches evaluated on Spider-Syn.

We compare the result of $\textnormal{RAT-SQL}_{\textnormal{\small{B}}}$ trained on Spider (SPR) as a baseline with other approaches. 
$\textnormal{RAT-SQL}_{\textnormal{\small{B}}}$ trained on Spider-Syn ($\textnormal{SPR}_{\textnormal{\small{SYN}}}$) obtains 11.7\% accuracy improvement when evaluated on Spider-Syn, while only suffers 1.9\% accuracy drop when evaluated on Spider.
Meanwhile, our adversarial training method based on BERT-Attack ($\textnormal{ADV}_{\textnormal{\small{BERT}}}$) improves the accuracy by 10.3\% on Spider-Syn.
We observe that $\textnormal{ADV}_{\textnormal{\small{BERT}}}$ performs much better than adversarial training based on GLOVE ($\textnormal{ADV}_{\textnormal{\small{GLOVE}}}$), and we provide more explanation in Section~\ref{sec:exp-adv}.
Both of our multiple annotation methods (ManualMAS and AutoMAS) improve the baseline model evaluated on Spider-Syn.
The performance of ManualMAS is better because the synonyms in ManualMAS are exactly the same as the synonym substitution in Spider-Syn. We discuss more results about multi-annotation selection in Section \ref{section:ablation-study}.

\begin{table}[t]
    \centering
    \resizebox{.99\columnwidth}{!}{
    \smallskip\begin{tabular}{lcc}
        \hline
       \bf Approach &  $\textbf{ADV}_{\bf GLOVE}$ &  $\textbf{ADV}_{\bf BERT}$ \\
        \hline \hline
        SPR  & 38.0\% &  48.8\% \\ 
        $\textnormal{SPR}_{\textnormal{\small{SYN}}}$  & 49.6\% &  54.9\% \\ 
        $\textnormal{SPR}_{\textnormal{\small{SPR\&SYN}}}$  & 47.7\% &  55.7\% \\ 
        \hline
        $\textnormal{ADV}_{\textnormal{\small{GLOVE}}}$  & 29.7\% &  33.8\% \\ 
        $\textnormal{ADV}_{\textnormal{\small{BERT}}}$  & 55.7\% & \bf 59.2\% \\ 
        \hline
        SPR + ManualMAS  & 34.2\% &  44.5\% \\ 
        SPR + AutoMAS  & \bf 61.2\% &  52.5\% \\ 
        \hline 
    \end{tabular}
    }
    \caption{Exact match accuracy on the worst-case development sets generated by $\textnormal{ADV}_{\textnormal{\small{GLOVE}}}$ and $\textnormal{ADV}_{\textnormal{\small{BERT}}}$. All approaches use the $\textnormal{RAT-SQL}_{\textnormal{\small{B}}}$ model.}\smallskip
    \label{table:attack-defense}
\end{table}

\begin{table*}[t]
    \centering
    \smallskip\begin{tabular}{lcccc}
        \hline
       \bf Approach & \bf Spider & \bf Spider-Syn & $\textbf{ADV}_{\bf GLOVE}$ &  $\textbf{ADV}_{\bf BERT}$ \\
        \hline \hline
        SPR  & \bf 69.7\% &  48.2\% & 38.0\% & 48.8\% \\ 
        SPR + ManualMAS  &  67.4\% & \bf  62.6\% & 34.2\% & 44.5\% \\ 
        SPR + AutoMAS &  68.7\% &  56.0\% & \bf  61.2\% & \bf 52.5\% \\ 
        \hline
        $\textnormal{SPR}_{\textnormal{\small{SYN}}}$  & \bf 67.8\% &  59.9\% & 49.6\% & \bf 54.9\% \\ 
        $\textnormal{SPR}_{\textnormal{\small{SYN}}}$ + ManualMAS   & 65.7\% &  \bf 62.9\% & 47.8\% & 52.1\% \\ 
        $\textnormal{SPR}_{\textnormal{\small{SYN}}}$ + AutoMAS  & 67.0\% &  61.7\% & \bf 63.3\% & 54.4\% \\ 
        \hline
        
        $\textnormal{SPR\&SPR}_{\textnormal{\small{SYN}}}$  & \bf 68.1\% &  58.0\% & 47.7\% & \bf 55.7\% \\ 
        $\textnormal{SPR\&SPR}_{\textnormal{\small{SYN}}}$ + ManualMAS   & 65.6\% &  \bf 59.5\% & 46.9\% & 51.7\% \\ 
        $\textnormal{SPR\&SPR}_{\textnormal{\small{SYN}}}$ + AutoMAS  & 66.8\% &  57.5\% & \bf 61.0\% & \bf 55.7\% \\ 
        \hline
        
        $\textnormal{ADV}_{\textnormal{\small{BERT}}}$  & \bf 68.7\% &  58.5\% & 55.7\% & \bf 59.2\% \\ 
        $\textnormal{ADV}_{\textnormal{\small{BERT}}}$ + ManualMAS  & 66.7\% & \bf  62.2\%  & 53.4\%  & 56.7\%\\ 
        $\textnormal{ADV}_{\textnormal{\small{BERT}}}$ + AutoMAS   & 67.5\% &  59.6\% & \bf 62.4\% & 58.0\% \\ 
        \hline 
    \end{tabular}
    \caption{Ablation study results using $\textnormal{RAT-SQL}_{\textnormal{\small{B}}}$.}\smallskip
    \label{table:Ablation-study}
\end{table*}

\subsection{Evaluation on Adversarial Attacks}
\label{sec:exp-adv}

Observing the dramatic performance drop on Spider-Syn, we then study the model robustness under worst-case attacks. 
We use the adversarial examples generation module in $\textnormal{ADV}_{\textnormal{\small{GLOVE}}}$ and $\textnormal{ADV}_{\textnormal{\small{BERT}}}$ to attack the $\textnormal{RAT-SQL}_{\textnormal{\small{B}}}$+SPR to generate two worst-case development sets.

Table \ref{table:attack-defense} presents the results on two worst-case development sets. 
The $\textnormal{ADV}_{\textnormal{\small{GLOVE}}}$ and $\textnormal{ADV}_{\textnormal{\small{BERT}}}$ attacks cause the accuracy of $\textnormal{RAT-SQL}_{\textnormal{\small{B}}}$+SPR to drop by 31.7\% and 20.9\%, respectively.
$\textnormal{RAT-SQL}_{\textnormal{\small{B}}}$+SPR+AutoMAS achieve the best performance on defending the $\textnormal{ADV}_{\textnormal{\small{GLOVE}}}$ attack. 
Because the annotations in AutoMAS cover the synonym substitutions generated by $\textnormal{ADV}_{\textnormal{\small{GLOVE}}}$. 
The relation between AutoMAS and $\textnormal{ADV}_{\textnormal{\small{GLOVE}}}$ is similar to that between ManualMAS and Spider-Syn.
Similarly, ManualMAS helps $\textnormal{RAT-SQL}_{\textnormal{\small{B}}}$+SPR get the best accuracy as shown in Table \ref{table:our-approaches}.

As to $\textnormal{ADV}_{\textnormal{\small{BERT}}}$ attack, $\textnormal{RAT-SQL}_{\textnormal{\small{B}}}$+$\textnormal{ADV}_{\textnormal{\small{BERT}}}$ outperforms other approaches.
This result is not surprising, because $\textnormal{RAT-SQL}_{\textnormal{\small{B}}}$+$\textnormal{ADV}_{\textnormal{\small{BERT}}}$ is trained based on defense $\textnormal{ADV}_{\textnormal{\small{BERT}}}$ attack. 
However, why does $\textnormal{RAT-SQL}_{\textnormal{\small{B}}}$+$\textnormal{ADV}_{\textnormal{\small{GLOVE}}}$ perform so poorly in defending $\textnormal{ADV}_{\textnormal{\small{GLOVE}}}$ attack? 

We conjecture that this is because the word embedding from BERT is based on the context: if you replace a word with a so-called synonym that is irrelevant to the context, BERT may give this synonym a vector with low similarity to the original.
In the first example of Table \ref{table:question-examples}, $\textnormal{ADV}_{\textnormal{\small{GLOVE}}}$ replaces the word `\emph{courses}' with `\emph{trajectory}'.
We observe that, based on the cosine similarity of BERT embedding, the schema item most similar to `\emph{trajectory}' changes from `\emph{courses}' to `\emph{grade conversion}'.
This problem does not appear in the Spider-Syn and $\textnormal{ADV}_{\textnormal{\small{BERT}}}$ examples, and some $\textnormal{ADV}_{\textnormal{\small{GLOVE}}}$ examples do not have this problem, such as the second example in Table \ref{table:question-examples}. 
Some examples reward the model for finding the schema item that is most similar to the question token, while others penalize this pattern, which causes the model to fail to learn. 
Thus the model with $\textnormal{ADV}_{\textnormal{\small{GLOVE}}}$ neither defends against $\textnormal{ADV}_{\textnormal{\small{GLOVE}}}$ attack nor even obtains good performance on the Spider.

\begin{table*}[t]
    \centering
    \scalebox{0.9}{
    \begin{tabular}{p{0.25\columnwidth}p{1.9\columnwidth}}
  
  Spider:& Which {\bf {courses}} are taught on {\bf {days}} MTW?\\\hline
  
  Spider-Syn: & Which {\bf \textcolor{red}{curriculum}} are taught on days MTW? \\\hline
  $\textnormal{ADV}_{\textnormal{\small{GLOVE}}}$: &  Which {\bf \textcolor{red}{trajectory}} are taught on {\bf \textcolor{red}{jour}} MTW ? \\\hline
  
  $\textnormal{ADV}_{\textnormal{\small{BERT}}}$:    &  Which {\bf \textcolor{red}{classes}} are taught on {\bf \textcolor{red}{times}} MTW ? \\\hline
    \\

    Spider:& Show the name and {\bf {phone}} for {\bf {customers}} with a {\bf {mailshot}} with {\bf {outcome}} code `No Response'\\\hline
  
    Spider-Syn: & Show the name and {\bf \textcolor{red}{telephone}} for {\bf \textcolor{red}{clients}} with a mailshot with outcome code `No Response'. \\\hline
    $\textnormal{ADV}_{\textnormal{\small{GLOVE}}}$: &  Show the name and {\bf \textcolor{red}{telephones}} for customers with a mailshot with outcome code `No Response'. \\\hline
    
    $\textnormal{ADV}_{\textnormal{\small{BERT}}}$:    &  Show the name and {\bf \textcolor{red}{telephone}} for customers with a {\bf \textcolor{red}{mailbox}} with {\bf \textcolor{red}{result}} code `No Response'. \\\hline

    \end{tabular}}
    \caption{Two questions in Spider with corresponding versions of Spider-Syn, $\textnormal{ADV}_{\textnormal{\small{GLOVE}}}$ and $\textnormal{ADV}_{\textnormal{\small{BERT}}}$. }\smallskip
    \label{table:question-examples}
  
    \end{table*}

\subsection{Ablation Study}
\label{section:ablation-study}
To analyze the individual contribution of our proposed techniques, we have run some additional experiments and show their results in Table \ref{table:Ablation-study}.
Specifically, we use 
$\textnormal{RAT-SQL}_{\textnormal{\small{B}}}$+SPR, $\textnormal{RAT-SQL}_{\textnormal{\small{B}}}$+$\textnormal{SPR}_{\textnormal{\small{SYN}}}$, $\textnormal{RAT-SQL}_{\textnormal{\small{B}}}$+$\textnormal{SPR}_{\textnormal{\small{SPR\&SYN}}}$, and $\textnormal{RAT-SQL}_{\textnormal{\small{B}}}$+$\textnormal{ADV}_{\textnormal{\small{BERT}}}$ as base models, then we apply different schema annotation methods to these model and evaluate their performance in different development sets. Note that all base models use the Spider original schema annotations.

First, for all base models, we found that MAS consistently improves the model performance when questions are modified by synonym substitution.
Specifically, when evaluating on Spider-Syn, using ManualMAS achieves the best performance, 
because the ManualMAS contains the synonym substitutions of Spider-Syn.
Meanwhile, when evaluating on worst-case adversarial attacks, AutoMAS mostly outperforms ManualMAS. Considering that the AutoMAS is automatically generated, AutoMAS would be a simple and efficient way to improve the robustness of text-to-SQL models. 

\subsection{Further Discussion on MAS}
ManualMAS utilizes the same synonym annotations on Spider-Syn, the same relationship as AutoMAS with $\textnormal{ADV}_{\textnormal{\small{GLOVE}}}$, and we design this mechanism to demonstrate the effectiveness of MAS in an ideal case.
By showing the superior performance of ManualMAS on Spider-Syn, we confirm that the failure of existing models on Spider-Syn is largely because they rely on the lexical correspondence, and MAS improves the performance by repairing the lexical link. Besides, MAS has the following advantages:

\begin{itemize}[leftmargin=*,noitemsep,topsep=0em]
    \item Compared to adversarial training, MAS does not need any additional training. Therefore, by including different annotations for MAS, the same pre-trained model could be applied to application scenarios with different requirements of robustness to synonym substitutions.
    \item MAS could also be combined with existing defenses, e.g., on adversarially trained models, as shown in our evaluation.
\end{itemize}

We add the evaluation on the combination of MAS with GNN and IRNet respectively, shown in Table~\ref{table:GNN+MAS}. The conclusions are similar to RAT-SQL: (1) MAS significantly improves the performance on Spider-Syn, and ManualMAS achieves the best performance. (2) AutoMAS also considerably improves the performance on adversarial attacks.

\begin{table*}[t]
    \centering
    \smallskip\begin{tabular}{lcccc}
        \hline
       \bf Approach & \bf Spider & \bf Spider-Syn & $\textbf{ADV}_{\bf GLOVE}$ &  $\textbf{ADV}_{\bf BERT}$ \\
        \hline \hline
        GNN  & \bf 48.5\% &  23.6\% & 25.4\% & 28.9\% \\ 
        GNN + ManualMAS  &  44.0\% & \bf  38.2\% & 22.9\% & 26.2\% \\ 
        GNN + AutoMAS &  44.0\% &  29.5\% & \bf  39.8\% & \bf 31.8\% \\ 
        \hline
        IRNet  & \bf 53.2\% &  28.4\% & 26.4\% &  29.0\% \\ 
        IRNet + ManualMAS   & 49.7\% &  \bf 39.3\% & 24.0\% & 27.2\% \\ 
        IRNet + AutoMAS  & 53.1\% &  35.1\% & \bf 44.3\% & \bf 35.6\% \\ 
        \hline

    \end{tabular}
    \caption{Evaluation on the combination of MAS with GNN and IRNet respectively.}\smallskip
    \label{table:GNN+MAS}
\end{table*}



\section{Related Work}

\paragraph{Text-to-SQL translation.} Text-to-SQL translation has been a long-standing challenge, and various benchmarks are constructed for this task~\cite{data-atis-geography-scholar,data-restaurants-original,data-restaurants-logic,data-restaurants,Li2014,data-sql-imdb-yelp,zhongSeq2SQL2017,Yu2018a}. In particular, most recent works aim to improve the performance on Spider benchmark~\cite{Yu2018a}, where models are required to synthesize SQL queries with complex structures, e.g., JOIN clauses and nested queries, and they need to generalize across databases of different domains.
Among various model architectures~\cite{Yu2018-SyntaxSQLNet,Bogin2019,Guo2019,Zhang2019,Bogin2019a,Wang2019}, latest state-of-the-art models have implemented a schema linking method, which is based on the exact lexical matching between the NL question and the table schema items~\cite{Guo2019,Bogin2019,Wang2019}. Schema linking is essential for these models, and causes a huge performance drop when removing it. Based on this observation, we investigate the robustness of such models to synonym substitution in this work.

\paragraph{Data augmentation for text-to-SQL models.} 
Existing works have proposed some data augmentation and adversarial training techniques to improve the performance of text-to-SQL models.
\citet{Xiong2019} propose an AugmentGAN model to generate samples in the target domain for data augmentation, so as to improve the cross-domain generalization. However, this approach only supports SQL queries executed on a single table, e.g., WikiSQL.
\citet{Li2019} propose to use data augmentation specialized for learning the spatial information in databases, which improves the performance on single-domain GeoQuery and Restaurants datasets.
Some recent works study data augmentation to improve the model performance on variants of existing SQL benchmarks. Specifically, \citet{Radhakrishnan2020} focus on search-style questions that are short and colloquial, and \citet{Zhu2020} study adversarial training to improve the adversarial robustness. However, both of them are based on WikiSQL.
\citet{Zeng2020} study the model robustness when the NL questions are untranslatable and ambiguous, where they construct a dataset of such questions based on the Spider benchmark, and perform data augmentation to detect confusing spans in the question.
On the contrary, our work investigate the robustness against synonym substitution for cross-domain text-to-SQL translation, supporting complex SQL structures.

\paragraph{Synonym substitution for other NLP problems.} 
The study of synonym substitution can be traced back to the 1970s \cite{waltz1978english,lehmann1972normalization}.
With the rise of machine learning, synonym substitution is widely used in NLP for data augment and adversarial attacks \cite{10.1145/3357384.3358040,wei-zou-2019-eda,ebrahimi-etal-2018-hotflip,alshemali-kalita-2020-generalization,ren-etal-2019-generating}. 
Many adversarial attacks based on synonym substitution have successfully compromised the performance of existing models \cite{alzantot2018generating,zhang-etal-2019-generating-fluent,ren-etal-2019-generating,jin2020bert}. 
Recently, \cite{Morris2020} integrate many above works into their TextAttack framework for ease of use.

\section{Conclusion}
We introduce Spider-Syn, a human-curated dataset based on the Spider benchmark for evaluating the robustness of text-to-SQL models for synonym substitution. 
We found that the performance of previous text-to-SQL models drop dramatically on Spider-Syn, as well as other adversarial attacks performing the synonym substitution.
We design two categories of approaches to improve the model robustness, i.e., multi-anotation selection and adversarial training, and demonstrate the effectiveness of our approaches.

\section*{Acknowledgements}
We would like to thank the anonymous reviewers for their helpful comments.
Matthew Purver is partially supported by the EPSRC under grant EP/S033564/1, and by the European Union's Horizon 2020 programme under grant agreements 769661 (SAAM, Supporting Active Ageing through Multimodal coaching) and 825153 (EMBEDDIA, Cross-Lingual Embeddings for Less-Represented Languages in European News Media). Xinyun Chen is supported by the Facebook Fellowship. The results of this publication reflect only the authors' views and the Commission is not responsible for any use that may be made of the information it contains.



\bibliographystyle{acl_natbib}
\bibliography{acl2021}

\begin{thebibliography}{35}
\expandafter\ifx\csname natexlab\endcsname\relax\def\natexlab#1{#1}\fi

\bibitem[{Alshemali and Kalita(2020)}]{alshemali-kalita-2020-generalization}
Basemah Alshemali and Jugal Kalita. 2020.
\newblock \href {https://doi.org/10.18653/v1/2020.deelio-1.3} {{Generalization
  to Mitigate Synonym Substitution Attacks}}.
\newblock In \emph{Proceedings of Deep Learning Inside Out (DeeLIO): The First
  Workshop on Knowledge Extraction and Integration for Deep Learning
  Architectures}, pages 20--28, Online. Association for Computational
  Linguistics.

\bibitem[{Alzantot et~al.(2018)Alzantot, Sharma, Elgohary, Ho, Srivastava, and
  Chang}]{alzantot2018generating}
Moustafa Alzantot, Yash Sharma, Ahmed Elgohary, Bo-Jhang Ho, Mani Srivastava,
  and Kai-Wei Chang. 2018.
\newblock Generating natural language adversarial examples.
\newblock In \emph{Proceedings of the 2018 Conference on Empirical Methods in
  Natural Language Processing}, pages 2890--2896.

\bibitem[{{Ana-Maria Popescu} et~al.(2003){Ana-Maria Popescu}, Etzioni, and
  Kautz}]{data-restaurants-original}
{Ana-Maria Popescu}, Oren Etzioni, and Henry Kautz. 2003.
\newblock \href {http://doi.acm.org/10.1145/604045.604070} {{Towards a Theory
  of Natural Language Interfaces to Databases}}.
\newblock In \emph{Proceedings of the 8th International Conference on
  Intelligent User Interfaces}, pages 149--157.

\bibitem[{Bogin et~al.(2019{\natexlab{a}})Bogin, Berant, and
  Gardner}]{Bogin2019}
Ben Bogin, Jonathan Berant, and Matt Gardner. 2019{\natexlab{a}}.
\newblock \href {https://doi.org/10.18653/v1/P19-1448} {Representing schema
  structure with graph neural networks for text-to-{SQL} parsing}.
\newblock In \emph{Proceedings of the 57th Annual Meeting of the Association
  for Computational Linguistics}, pages 4560--4565, Florence, Italy.
  Association for Computational Linguistics.

\bibitem[{Bogin et~al.(2019{\natexlab{b}})Bogin, Gardner, and
  Berant}]{Bogin2019a}
Ben Bogin, Matt Gardner, and Jonathan Berant. 2019{\natexlab{b}}.
\newblock \href {https://doi.org/10.18653/v1/D19-1378} {{Global Reasoning over
  Database Structures for Text-to-{SQL} Parsing}}.
\newblock In \emph{Proceedings of the 2019 Conference on Empirical Methods in
  Natural Language Processing and the 9th International Joint Conference on
  Natural Language Processing (EMNLP-IJCNLP)}, pages 3659--3664, Hong Kong,
  China. Association for Computational Linguistics.

\bibitem[{Ebrahimi et~al.(2018)Ebrahimi, Rao, Lowd, and
  Dou}]{ebrahimi-etal-2018-hotflip}
Javid Ebrahimi, Anyi Rao, Daniel Lowd, and Dejing Dou. 2018.
\newblock \href {https://doi.org/10.18653/v1/P18-2006} {{H}ot{F}lip: White-box
  adversarial examples for text classification}.
\newblock In \emph{Proceedings of the 56th Annual Meeting of the Association
  for Computational Linguistics (Volume 2: Short Papers)}, pages 31--36,
  Melbourne, Australia. Association for Computational Linguistics.

\bibitem[{Giordani and Moschitti(2012)}]{data-restaurants}
Alessandra Giordani and Alessandro Moschitti. 2012.
\newblock \href {https://doi.org/10.1007/978-3-642-45260-4{\_}5} {{Automatic
  Generation and Reranking of SQL-derived Answers to NL Questions}}.
\newblock In \emph{Proceedings of the Second International Conference on
  Trustworthy Eternal Systems via Evolving Software, Data and Knowledge}, pages
  59--76.

\bibitem[{Guo et~al.(2019)Guo, Zhan, Gao, Xiao, Lou, Liu, and Zhang}]{Guo2019}
Jiaqi Guo, Zecheng Zhan, Yan Gao, Yan Xiao, Jian-Guang Lou, Ting Liu, and
  Dongmei Zhang. 2019.
\newblock \href {https://doi.org/10.18653/v1/P19-1444} {{Towards Complex
  Text-to-{SQL} in Cross-Domain Database with Intermediate Representation}}.
\newblock In \emph{Proceedings of the 57th Annual Meeting of the Association
  for Computational Linguistics}, pages 4524--4535, Florence, Italy.
  Association for Computational Linguistics.

\bibitem[{Iyer et~al.(2017)Iyer, Konstas, Cheung, Krishnamurthy, and
  Zettlemoyer}]{data-atis-geography-scholar}
Srinivasan Iyer, Ioannis Konstas, Alvin Cheung, Jayant Krishnamurthy, and Luke
  Zettlemoyer. 2017.
\newblock \href {https://doi.org/10.18653/v1/P17-1089} {{Learning a Neural
  Semantic Parser from User Feedback}}.
\newblock In \emph{Proceedings of the 55th Annual Meeting of the Association
  for Computational Linguistics (Volume 1: Long Papers)}, pages 963--973.

\bibitem[{Jin et~al.(2020)Jin, Jin, Zhou, and Szolovits}]{jin2020bert}
Di~Jin, Zhijing Jin, Joey~Tianyi Zhou, and Peter Szolovits. 2020.
\newblock Is bert really robust? a strong baseline for natural language attack
  on text classification and entailment.
\newblock In \emph{Proceedings of the AAAI conference on artificial
  intelligence}, volume~34, pages 8018--8025.

\bibitem[{Lehmann and Stachowitz(1972)}]{lehmann1972normalization}
Winfred~Philipp Lehmann and RA~Stachowitz. 1972.
\newblock Normalization of natural language for information retrieval.
\newblock Technical report, TEXAS UNIV AUSTIN LINGUISTICS RESEARCH CENTER.

\bibitem[{Li and Jagadish(2014)}]{Li2014}
Fei Li and H.~V. Jagadish. 2014.
\newblock \href {https://doi.org/10.14778/2735461.2735468} {{Constructing an
  interactive natural language interface for relational databases}}.
\newblock \emph{Proceedings of the VLDB Endowment}, 8(1):73--84.

\bibitem[{Li et~al.(2019)Li, Wang, Ku, Tian, and Wang}]{Li2019}
Jingjing Li, Wenlu Wang, Wei~Shinn Ku, Yingtao Tian, and Haixun Wang. 2019.
\newblock \href {https://doi.org/10.1145/3347146.3359069} {{SpatialNLI: A
  spatial domain natural language interface to databases using spatial
  comprehension}}.
\newblock In \emph{GIS: Proceedings of the ACM International Symposium on
  Advances in Geographic Information Systems}, pages 339--348, New York, NY,
  USA. Association for Computing Machinery.

\bibitem[{Li et~al.(2020)Li, Ma, Guo, Xue, and Qiu}]{Li2020}
Linyang Li, Ruotian Ma, Qipeng Guo, Xiangyang Xue, and Xipeng Qiu. 2020.
\newblock \href {https://doi.org/10.18653/v1/2020.emnlp-main.500}
  {{BERT-ATTACK: Adversarial Attack Against BERT Using BERT}}.
\newblock In \emph{Proceedings of the 2020 Conference on Empirical Methods in
  Natural Language Processing (EMNLP)}, pages 6193--6202, Stroudsburg, PA, USA.
  Association for Computational Linguistics.

\bibitem[{Madry et~al.(2018)Madry, Makelov, Schmidt, Tsipras, and
  Vladu}]{aleks2017deep}
Aleksander Madry, Aleksandar Makelov, Ludwig Schmidt, Dimitris Tsipras, and
  Adrian Vladu. 2018.
\newblock Towards deep learning models resistant to adversarial attacks.
\newblock In \emph{International Conference on Learning Representations}.

\bibitem[{Morris et~al.(2020)Morris, Lifland, Yoo, Grigsby, Jin, and
  Qi}]{Morris2020}
John Morris, Eli Lifland, Jin~Yong Yoo, Jake Grigsby, Di~Jin, and Yanjun Qi.
  2020.
\newblock \href {https://doi.org/10.18653/v1/2020.emnlp-demos.16} {{TextAttack:
  A Framework for Adversarial Attacks, Data Augmentation, and Adversarial
  Training in NLP}}.
\newblock In \emph{Proceedings of the 2020 Conference on Empirical Methods in
  Natural Language Processing: System Demonstrations}, pages 119--126,
  Stroudsburg, PA, USA. Association for Computational Linguistics.

\bibitem[{Mrk{\v{s}}i{\'c} et~al.(2016)Mrk{\v{s}}i{\'c}, {\'O}~S{\'e}aghdha,
  Thomson, Ga{\v{s}}i{\'c}, Rojas-Barahona, Su, Vandyke, Wen, and
  Young}]{mrksic-etal-2016-counter}
Nikola Mrk{\v{s}}i{\'c}, Diarmuid {\'O}~S{\'e}aghdha, Blaise Thomson, Milica
  Ga{\v{s}}i{\'c}, Lina~M. Rojas-Barahona, Pei-Hao Su, David Vandyke,
  Tsung-Hsien Wen, and Steve Young. 2016.
\newblock \href {https://doi.org/10.18653/v1/N16-1018} {Counter-fitting word
  vectors to linguistic constraints}.
\newblock In \emph{Proceedings of the 2016 Conference of the North {A}merican
  Chapter of the Association for Computational Linguistics: Human Language
  Technologies}, pages 142--148, San Diego, California. Association for
  Computational Linguistics.

\bibitem[{Pennington et~al.(2014)Pennington, Socher, and
  Manning}]{pennington-etal-2014-glove}
Jeffrey Pennington, Richard Socher, and Christopher Manning. 2014.
\newblock \href {https://doi.org/10.3115/v1/D14-1162} {{G}love: Global vectors
  for word representation}.
\newblock In \emph{Proceedings of the 2014 Conference on Empirical Methods in
  Natural Language Processing ({EMNLP})}, pages 1532--1543, Doha, Qatar.
  Association for Computational Linguistics.

\bibitem[{Radhakrishnan et~al.(2020)Radhakrishnan, Srikantan, and
  Lin}]{Radhakrishnan2020}
Karthik Radhakrishnan, Arvind Srikantan, and Xi~Victoria Lin. 2020.
\newblock \href {http://arxiv.org/abs/2010.09927} {{ColloQL: Robust
  Cross-Domain Text-to-SQL Over Search Queries}}.

\bibitem[{Ren et~al.(2019)Ren, Deng, He, and Che}]{ren-etal-2019-generating}
Shuhuai Ren, Yihe Deng, Kun He, and Wanxiang Che. 2019.
\newblock \href {https://doi.org/10.18653/v1/P19-1103} {{Generating Natural
  Language Adversarial Examples through Probability Weighted Word Saliency}}.
\newblock In \emph{Proceedings of the 57th Annual Meeting of the Association
  for Computational Linguistics}, pages 1085--1097, Florence, Italy.
  Association for Computational Linguistics.

\bibitem[{Rizos et~al.(2019)Rizos, Hemker, and
  Schuller}]{10.1145/3357384.3358040}
Georgios Rizos, Konstantin Hemker, and Bj\"{o}rn Schuller. 2019.
\newblock \href {https://doi.org/10.1145/3357384.3358040} {Augment to prevent:
  Short-text data augmentation in deep learning for hate-speech
  classification}.
\newblock In \emph{Proceedings of the 28th ACM International Conference on
  Information and Knowledge Management}, CIKM '19, page 991–1000, New York,
  NY, USA. Association for Computing Machinery.

\bibitem[{Tang and Mooney(2000)}]{data-restaurants-logic}
Lappoon~R Tang and Raymond~J Mooney. 2000.
\newblock \href {http://www.aclweb.org/anthology/W00-1317} {{Automated
  Construction of Database Interfaces: Intergrating Statistical and Relational
  Learning for Semantic Parsing}}.
\newblock In \emph{2000 Joint SIGDAT Conference on Empirical Methods in Natural
  Language Processing and Very Large Corpora}, pages 133--141.

\bibitem[{Waltz(1978)}]{waltz1978english}
David~L Waltz. 1978.
\newblock An english language question answering system for a large relational
  database.
\newblock \emph{Communications of the ACM}, 21(7):526--539.

\bibitem[{Wang et~al.(2020)Wang, Shin, Liu, Polozov, and Richardson}]{Wang2019}
Bailin Wang, Richard Shin, Xiaodong Liu, Oleksandr Polozov, and Matthew
  Richardson. 2020.
\newblock \href {https://doi.org/10.18653/v1/2020.acl-main.677} {{{RAT-SQL}:
  Relation-Aware Schema Encoding and Linking for Text-to-{SQL} Parsers}}.
\newblock In \emph{Proceedings of the 58th Annual Meeting of the Association
  for Computational Linguistics}, pages 7567--7578, Online. Association for
  Computational Linguistics.

\bibitem[{Wei and Zou(2019)}]{wei-zou-2019-eda}
Jason Wei and Kai Zou. 2019.
\newblock \href {https://doi.org/10.18653/v1/D19-1670} {{EDA}: Easy data
  augmentation techniques for boosting performance on text classification
  tasks}.
\newblock In \emph{Proceedings of the 2019 Conference on Empirical Methods in
  Natural Language Processing and the 9th International Joint Conference on
  Natural Language Processing (EMNLP-IJCNLP)}, pages 6382--6388, Hong Kong,
  China. Association for Computational Linguistics.

\bibitem[{Xiong and Sun(2019)}]{Xiong2019}
Hongvu Xiong and Ruixiao Sun. 2019.
\newblock \href {https://doi.org/10.1109/ICOSC.2019.8665499} {{Transferable
  Natural Language Interface to Structured Queries Aided by Adversarial
  Generation}}.
\newblock In \emph{2019 IEEE 13th International Conference on Semantic
  Computing (ICSC)}, pages 255--262. IEEE.

\bibitem[{Yaghmazadeh et~al.(2017)Yaghmazadeh, Wang, Dillig, and
  Dillig}]{data-sql-imdb-yelp}
Navid Yaghmazadeh, Yuepeng Wang, Isil Dillig, and Thomas Dillig. 2017.
\newblock \href {http://doi.org/10.1145/3133887} {{SQLizer: Query Synthesis
  from Natural Language}}.
\newblock In \emph{International Conference on Object-Oriented Programming,
  Systems, Languages, and Applications, ACM}, pages 63:1----63:26.

\bibitem[{Yu et~al.(2018{\natexlab{a}})Yu, Yasunaga, Yang, Zhang, Wang, Li, and
  Radev}]{Yu2018-SyntaxSQLNet}
Tao Yu, Michihiro Yasunaga, Kai Yang, Rui Zhang, Dongxu Wang, Zifan Li, and
  Dragomir Radev. 2018{\natexlab{a}}.
\newblock \href {https://doi.org/10.18653/v1/D18-1193} {{S}yntax{SQLN}et:
  Syntax tree networks for complex and cross-domain text-to-{SQL} task}.
\newblock In \emph{Proceedings of the 2018 Conference on Empirical Methods in
  Natural Language Processing}, pages 1653--1663, Brussels, Belgium.
  Association for Computational Linguistics.

\bibitem[{Yu et~al.(2018{\natexlab{b}})Yu, Zhang, Yang, Yasunaga, Wang, Li, Ma,
  Li, Yao, Roman, Zhang, and Radev}]{Yu2018a}
Tao Yu, Rui Zhang, Kai Yang, Michihiro Yasunaga, Dongxu Wang, Zifan Li, James
  Ma, Irene Li, Qingning Yao, Shanelle Roman, Zilin Zhang, and Dragomir Radev.
  2018{\natexlab{b}}.
\newblock \href {https://doi.org/10.18653/v1/D18-1425} {{S}pider: A large-scale
  human-labeled dataset for complex and cross-domain semantic parsing and
  text-to-{SQL} task}.
\newblock In \emph{Proceedings of the 2018 Conference on Empirical Methods in
  Natural Language Processing}, pages 3911--3921, Brussels, Belgium.
  Association for Computational Linguistics.

\bibitem[{Zelle and Mooney(1996)}]{data-geography-original}
John~M Zelle and Raymond~J Mooney. 1996.
\newblock \href {http://dl.acm.org/citation.cfm?id=1864519.1864543} {{Learning
  to Parse Database Queries Using Inductive Logic Programming}}.
\newblock In \emph{Proceedings of the Thirteenth National Conference on
  Artificial Intelligence - Volume 2}, pages 1050--1055.

\bibitem[{Zeng et~al.(2020)Zeng, Lin, Hoi, Socher, Xiong, Lyu, and
  King}]{Zeng2020}
Jichuan Zeng, Xi~Victoria Lin, Steven~C.H. Hoi, Richard Socher, Caiming Xiong,
  Michael Lyu, and Irwin King. 2020.
\newblock \href {https://doi.org/10.18653/v1/2020.acl-demos.24} {{Photon: A
  Robust Cross-Domain Text-to-SQL System}}.
\newblock In \emph{Proceedings of the 58th Annual Meeting of the Association
  for Computational Linguistics: System Demonstrations}, pages 204--214,
  Stroudsburg, PA, USA. Association for Computational Linguistics.

\bibitem[{Zhang et~al.(2019{\natexlab{a}})Zhang, Zhou, Miao, and
  Li}]{zhang-etal-2019-generating-fluent}
Huangzhao Zhang, Hao Zhou, Ning Miao, and Lei Li. 2019{\natexlab{a}}.
\newblock \href {https://doi.org/10.18653/v1/P19-1559} {Generating fluent
  adversarial examples for natural languages}.
\newblock In \emph{Proceedings of the 57th Annual Meeting of the Association
  for Computational Linguistics}, pages 5564--5569, Florence, Italy.
  Association for Computational Linguistics.

\bibitem[{Zhang et~al.(2019{\natexlab{b}})Zhang, Yu, Er, Shim, Xue, Lin, Shi,
  Xiong, Socher, and Radev}]{Zhang2019}
Rui Zhang, Tao Yu, He~Yang Er, Sungrok Shim, Eric Xue, Xi~Victoria Lin, Tianze
  Shi, Caiming Xiong, Richard Socher, and Dragomir Radev. 2019{\natexlab{b}}.
\newblock \href {http://arxiv.org/abs/1909.00786} {{Editing-Based SQL Query
  Generation for Cross-Domain Context-Dependent Questions}}.

\bibitem[{Zhong et~al.(2017)Zhong, Xiong, and Socher}]{zhongSeq2SQL2017}
Victor Zhong, Caiming Xiong, and Richard Socher. 2017.
\newblock {Seq2SQL: Generating Structured Queries from Natural Language using
  Reinforcement Learning}.
\newblock \emph{CoRR}, abs/1709.0.

\bibitem[{Zhu et~al.(2020)Zhu, Zhou, and Xia}]{Zhu2020}
Yi~Zhu, Yiwei Zhou, and Menglin Xia. 2020.
\newblock \href {http://arxiv.org/abs/2005.12696} {{Generating Semantically
  Valid Adversarial Questions for TableQA}}.

\end{thebibliography}


\end{document}